\documentclass[journal,twoside,web]{ieeecolor2}
\usepackage{generic}
\usepackage{cite}
\usepackage{amsmath,amssymb,amsfonts}
\usepackage{algorithmic}
\usepackage{graphicx}
\usepackage{amsmath,amsfonts}
\usepackage{graphicx}
\usepackage{subcaption}
\usepackage{booktabs}
\usepackage{multirow}
\usepackage{tabularx}
\usepackage{colortbl}
\usepackage{wrapfig}
\usepackage{textcomp}
\usepackage{arydshln}
\newcommand{\spacedhdashline}{%
    \cdashline{3-17}%
    \noalign{\vspace{0.3mm}}%
}

\def\BibTeX{{\rm B\kern-.05em{\sc i\kern-.025em b}\kern-.08em
    T\kern-.1667em\lower.7ex\hbox{E}\kern-.125emX}}
\begin{document}
\title{Personalized Control for Lower Limb Prosthesis Using Kolmogorov-Arnold Networks}
\author{Seyed Mojtaba Mohasel, Alireza Afzal Aghaei, Corey Pew
\thanks{Seyed Mojtaba Mohasel is with the Department of Mechanical and Industrial Engineering, Montana State University, Montana, USA (e-mail: Seyedmojtabamohasel@montana.edu).}
\thanks{Alireza Afzal Aghaei is an Independent Researcher, Isfahan, Iran (e-mail: alirezaafzalaghaei@gmail.com).}
\thanks{Corey Pew is with the Department of Mechanical and Industrial Engineering, Montana State University, Montana, USA (e-mail: corey.pew@montana.edu).}
}

\maketitle

\begin{abstract}

\textit{Objective:} This paper investigates the potential of learnable activation functions in Kolmogorov-Arnold Networks (KANs) for personalized control in a lower-limb prosthesis. In addition, user-specific vs. pooled training data is evaluated to improve machine learning (ML) and Deep Learning (DL) performance for turn intent prediction.

\textit{Method:} Inertial measurement unit (IMU) data from the shank were collected from five individuals with lower-limb amputation performing turning tasks in a laboratory setting. Ability to classify an upcoming turn was evaluated for Multilayer
Perceptron (MLP), Kolmogorov-Arnold Network (KAN), convolutional neural network (CNN), and fractional Kolmogorov-Arnold Networks (FKAN). The comparison of MLP and KAN (for ML models) and FKAN and CNN (for DL models) assessed the effectiveness of learnable activation functions. Models were trained separately on user-specific and pooled data to evaluate the impact of training data on their performance.

\textit{Results:} Learnable activation functions in KAN and FKAN did not yield significant improvement compared to MLP and CNN, respectively. Training on user-specific data yielded superior results compared to pooled data for ML models ($p < 0.05$). In contrast, no significant difference was observed between user-specific and pooled training for DL models. 

\textit{Significance:} These findings suggest that learnable activation functions may demonstrate distinct advantages in datasets involving more complex tasks and larger volumes. In addition, pooled training showed comparable performance to user-specific training in DL models, indicating that model training for prosthesis control can utilize data from multiple participants.

\end{abstract}


\section{Introduction} 
The purpose of a prosthesis is to replicate the biomechanical function of an intact limb \cite{lawson2014robotic}. Prosthesis components are often optimized for straight walking but can be too rigid during torsional movements, causing discomfort during turning, twisting, and pivoting \cite{hafner2002transtibial, rietman2002gait}. Consequently, the user may limit their activities, resulting in a decreased quality of life. 

\begin{wrapfigure}{r}{0.40\linewidth} 
    \centering
    \includegraphics[width=0.9\linewidth]{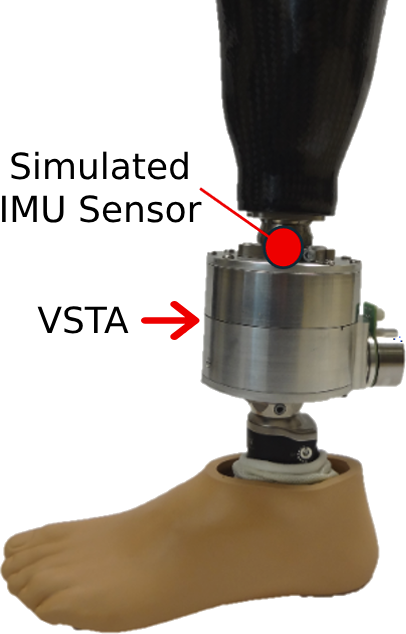}
    \caption{VSTA positioned below the socket provided variable torsional stiffness. Simulated IMU at proximal end of VSTA.}
    \label{fig:vsta}
\end{wrapfigure}

The Variable Stiffness Torsion Adapter (VSTA, Figure \ref{fig:vsta}) allows for semi-active stiffness variation in the transverse plane \cite{pew2015design}. The VSTA adjusts electromechanically in real time to optimize transverse stiffness, reducing biomechanical stress on the residual limb and joints \cite{pew2017pilot}. Modulation of the VSTA is limited to the swing phase of gait, when the device is unloaded, allowing approximately 300 milliseconds for gait detection and control action \cite{hudgins1993new}. This creates a challenge for a controller to rapidly detect changes in walking modes (straight vs. turning) with only an onboard microcontroller. 

Human activity recognition systems often rely on wearable sensors to collect input data and utilize machine learning (ML) techniques for activity classification. 
Electromyography (EMG)\cite{chen2013locomotion, huang2008strategy} and inertial measurement units (IMU) are commonly used sensors, however EMG signals are less robust, require skin contact, more processing power, and are subject to motion artifact \cite{pew2017turn}. IMU sensors are frequently used for classifying different actions of daily living \cite{young2014analysis,young2013training,zhuang2019design, yang2018wearable, ayman2019efficient}, however, limited research focuses on real-time detection of turns with IMUs\cite{golyski2017computational,novak2014toward,pew2017turn}.

A previous turn intent classifier for the VSTA utilized several classifiers, including a Support Vector Machine (SVM), K-Nearest Neighbor (KNN), and an ensemble model \cite{pew2017turn}. The SVM was identified as the top-performing model, achieving a macro average F1-score of 89\%  on individuals with lower-limb amputation. In addition, the study found that training the model on individual-specific data yielded higher macro average F1-score performance (89\%) compared to pooled (80\%) data training that combined data from multiple individuals  \cite{pew2017turn}. While initially acceptable for control, the previously proposed controllers require improvement to enhance user safety and comfort and would benefit from newer ML methods prior to a device like the VSTA being used commercially.

A macro-average F1 score improves when the ML model performs well across all classes. However, turning occurs less frequently compared to straight walking, leading to a class imbalance \cite{kaur2019systematic} and must be taken into account with an appropriate ML model
\cite{mohasel2025robust, ali2019imbalance, johnson2019survey}. Furthermore, improving data representation through feature extraction has the potential to enhance ML performance. ML models appropriate for the VSTA can be broadly organized into two groups \cite{wang2019deep}: 

\begin{enumerate}
    \item \textbf{Conventional Methods} often require human intervention for feature extraction  \cite{wang2019deep}. 
Examples previously used for prosthesis control include linear discriminant analysis
\cite{hargrove2013non}, quadratic discriminant analysis  \cite{ha2010volitional}, and SVM \cite{pew2017turn,huang2011continuous,hu2018fusion}. 
    The limitations of conventional methods are that the extracted features are constrained by human domain knowledge \cite{bengio2013deep}, and the generated features are often shallow \cite{yang2015deep,wang2019deep}.
    \item \textbf{Deep Learning (DL) Methods} generally have automatic feature extraction that eliminates some reliance on human expertise. In addition, DL can generate complex features, making it particularly suitable for challenging activity recognition tasks \cite{wang2019deep}. DL models that consider the temporal dependency of IMU sensor data include recurrent neural networks (RNNs) and convolutional neural networks (CNNs). While RNNs process sensor data sequentially, CNNs support parallel processing, which may reduce inference speed for prediction. This aspect of CNNs is particularly suitable for real-time intent recognition \cite{lu2020deep,su2019cnn,nazari2022comparison}.

\end{enumerate}

DL models utilize activation functions (mathematical operation applied to the weighted sum of a neuron’s inputs) to determining a neuron’s excitation and output. Traditionally, activation functions are predominantly static and non-linear. Recently, Kolmogorov-Arnold Networks (KAN) \cite{liu2024kan, liu2024kan2} have been developed that utilize a learnable B-spline activation function, providing the potential to further enhance DL performance. 

The learnable activation function of KAN provides flexibility for pattern recognition in activity classification, however, KAN research is still in its early stages \cite{aghaei2024fkan, al2024tcnn, vaca2024kolmogorov, livieris2024c}. Few studies have investigated KAN with IMU sensor data for human activity recognition tasks \cite{liu2024initial, yang2024adaptive}. Therefore, the potential of the learnable activation function in KAN for prosthesis control remains underexplored (Knowledge Gap 1).

Another approach to improving model performance is through appropriate data selection for model training. Utilizing the data from a single individual participant for model training has shown enhanced performance for conventional models \cite{pew2017turn}. However, DL performance improves with exposure to larger datasets \cite{bengio2017deep}. Including data from multiple participants can act as a double-edged sword for DL: more diverse data is beneficial for parameter tuning, but it may also reduce the emphasis on the unique characteristics of the target individual. Characteristics such as height, weight, age \cite{elbaz2018gait}, physical abilities, obesity, and foot arch height can affect gait \cite{kim2021effects}. It remains unclear whether DL models should prioritize individual-specific data during training for a given individual \cite{pew2017turn}, or if pooled data leads to higher performance (Knowledge Gap 2).

This research aims to develop an improved controller for the VSTA using KAN with appropriate data selection for improved performance. Our hypotheses addressing the identified knowledge gaps are as follows:

\begin{itemize}
    \item \textbf{Hypothesis 1}: Models with learnable activation functions will achieve a higher macro average F1 score for the VSTA controller than those with static activation functions.
    \item \textbf{Hypothesis 2}: KANs trained on individual-specific data will achieve higher performance (higher macro average F1 score) for the VSTA controller for each individual compared to KANs trained on pooled data from multiple individuals and tested on a specific individual.
    
\end{itemize}

\section{Methods}
\subsection{Data Properties}
\subsubsection{Data Collection}

An experimental protocol was previously conducted to provide straight walking and turning data and consisted of five male, unilateral, transtibial individuals with amputation as participants (Table \ref{tab:data-demographic}) \cite{pew2017pilot}. All participants provided informed consent to the Institutional Review Board-approved protocol, wore their as-prescribed suspension equipped with the VSTA, and a Vari-Flex Low Profile foot (Ossur, Reykjavik, ISL) to compensate for the additional height from the VSTA (Figure \ref{fig:vsta}).  A certified prosthetist aligned and adjusted the VSTA for each participant. Three stiffness settings were considered for VSTA testing: compliant (${0.30 \text{ Nm}}/{^\circ}$), intermediate (${0.57 \text{ Nm}}/{^\circ}$), and stiff (${0.91 \text{ Nm}}/{^\circ}$).

\begin{figure}[ht]
    \centering
    \includegraphics[width=0.8\linewidth]{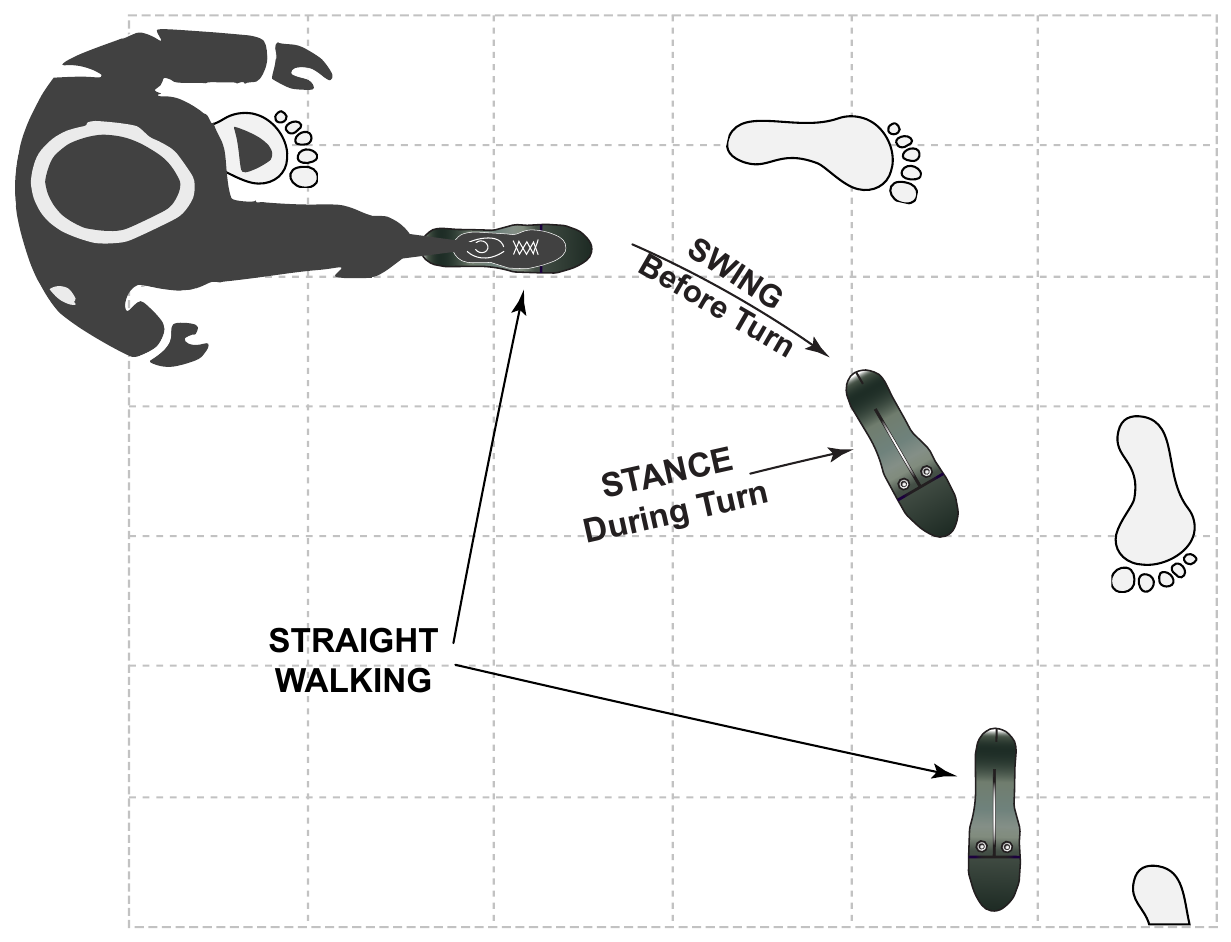}
    \caption{Illustration of a single-step 90° spin turn, showing the planted foot on the inside of the turn. The right leg represents the prosthesis and the left leg represents the intact limb without sensor.}
    \label{fig:total-prediction}
\end{figure}

Participants were instructed to perform three types of turns and straight walking at their self-selected pace (ranging from $1.29$ to $1.47$ m/s). Turn types included:
a) single-step 90° spin turns, where the planted foot was on the inside of the turn (Figure \ref{fig:total-prediction}); b) 90° step turns, where the planted foot was on the outside of the turn \cite{glaister2007video} \cite{pew2017pilot}; and  c) 180° turns with a single pivot foot.  In addition, the L-Test of functional mobility was performed \cite{deathe2005test}.

Each participant completed three to five trials per setting for each activity. Gait kinematics were recorded by a 12-camera Vicon motion capture system (Vicon, Centennial, CO) and a Plug-in-Gait marker model (120 Hz).

\begin{table}[ht]
\caption{Demographic and clinical characteristics of five individuals with trauma-related lower limb amputations.}
\resizebox{\columnwidth}{!}{%
\begin{tabular}{@{}ccccccc@{}}
\toprule
\textbf{Subject} & \textbf{Age} & \begin{tabular}[c]{@{}c@{}}\textbf{Weight}\\ (kg)\end{tabular} & \begin{tabular}[c]{@{}c@{}}\textbf{Height}\\ (m)\end{tabular} & \begin{tabular}[c]{@{}c@{}}\textbf{Amp.}\\ \textbf{Side}\end{tabular} & \begin{tabular}[c]{@{}c@{}}\textbf{Years}\\\textbf{Since Amp.}\end{tabular} & \begin{tabular}[c]{@{}c@{}}\textbf{Walking}\\ \textbf{Speed} (m/s)\end{tabular} \\ \midrule
A01 & 35 & 73.9 & 1.76 & Left & 11 & 1.32 \\
A02 & 47 & 84.8 & 1.77 & Left & 21 & 1.34 \\
A03 & 66 & 93.0 & 1.85 & Right & 48 & 1.47 \\
A04 & 47 & 98.4 & 1.89 & Right & 7 & 1.34 \\
A05 & 69 & 93.0 & 1.85 & Left & 42 & 1.29 \\ \bottomrule
\end{tabular}%
}
\label{tab:data-demographic}
\end{table}

\subsubsection{Data Post-Processing}

Raw Vicon position data, was processed in Visual 3D (HAS-Motion, Ontario, CA) to emulate IMU data from the shank. The simulated IMU data (Figure 1, considered proximal to the VSTA), generated in Visual 3D included three-axis translational velocity and three-axis angular velocity from the upper segment of the VSTA joint. Translational velocity data was then used to calculate segment acceleration in MATLAB (Mathworks, Natick, MA), applying a moving average filter with a seven-point window to smooth the signal. All data was referenced to the upper shank segment coordinate system to simulate the movement of an IMU within the prosthesis shank.
The stiffness setting to minimize peak transverse plane moments was independent of turn types \cite{pew2017pilot}. In addition, stiffness setting had no effect on the time to complete the L-Test of functional mobility, allowing all turning types with different stiffness levels to be combined into a single dataset \cite{pew2017pilot}. 

\subsubsection{Data characteristics}

Simulated IMU data consisting of linear acceleration (accelerometer) and angular velocity (gyroscope) in the X, Y, and Z directions were the inputs considered for all ML models. All data were labeled as follows: 

1) Straight walking (SW)

2) Pre-Turn, the swing phase immediately preceding the turn (SP):  The time from toe-off to the heel strike at the turn apex.

3) Turning, the stance phase at the turn apex (ST): Time from heel strike to toe-off during at the turn apex.

\subsubsection{Data Division}

Data for each participant is split based on total trials \cite{pew2017turn}, following the 80/20 split \cite{joseph2022split}. While stiffness setting had no measurable effect on turning kinematics, the test data was split to randomly include one trial from each turn type at each stiffness setting, for a total of nine turning events (three different turns with three different stiffness) to ensure no bias due to variable stiffness. Three trials of straight walking were also included to be representative of all actions. The remaining collected data for the specific individual constituted 80 percent of the data volume and was used as the training set.

\subsubsection{Data Segmentation}
Time varying data (acceleration and angular velocity) were grouped into overlapping, sliding windows \cite{banos2014window} of 10, 20, or 30 samples (83, 167, and 250 milliseconds) with 50\% overlap. This selection complies with the maximum prediction time constraint of 300 milliseconds (36 samples at 120 Hz). Window size was considered a hyperparameter, and Bayesian optimization (described later) selected the optimal window size along with other hyperparameters by evaluating model performance on validation data.

\subsection{ML Model Selection}

We selected a KAN and a Multilayer Perceptron (MLP) \cite{taud2017multilayer} as representatives of conventional methods to examine the impact of learnable activation functions compared to static activation functions in neural networks. Differences between MLP and KAN models are presented in Table \ref{tab:mlp_kan_comparison} \cite{liu2024kan}. In addition, we investigated learnable activation functions further by incorporating a Fractional Kolmogorov–Arnold Network (FKAN) \cite{aghaei2024fkan}, a deep learning variant of KAN, alongside a CNN as a representative of deep learning with automatic feature extraction. Both FKAN and CNN are capable of time series automatic feature extraction, while KAN and MLP are not.
 
\begin{table}[ht]
\centering
\caption{Comparison of MLP and KAN}

\setlength\extrarowheight{10pt}
\begin{tabular}{@{}lll@{}}
\toprule
\textbf{Aspect} & \textbf{MLP} & \textbf{KAN} \\ \midrule
\textbf{Theory} & Universal Approximation & \setlength\extrarowheight{0pt}\begin{tabular}[c]{@{}l@{}}Kolmogorov-Arnold\\ Representation\end{tabular}\setlength\extrarowheight{10pt} \\
\textbf{Activation Functions} & Fixed at nodes & Learnable on edges \\
\textbf{Shallow Structure} & \setlength\extrarowheight{0pt}\begin{tabular}[c]{@{}l@{}}Weights on edges\\ fixed activations at nodes\end{tabular}\setlength\extrarowheight{10pt} & \setlength\extrarowheight{0pt}\begin{tabular}[c]{@{}l@{}}Activations on edges\\ nodes sum operations\end{tabular}\setlength\extrarowheight{10pt} \\
\textbf{Deep Structure} & \setlength\extrarowheight{0pt}\begin{tabular}[c]{@{}l@{}}Linear transformations\\ fixed activations\end{tabular}\setlength\extrarowheight{10pt} & \setlength\extrarowheight{0pt}\begin{tabular}[c]{@{}l@{}}Layers with learnable\\ nonlinear transforms\end{tabular}\setlength\extrarowheight{10pt} \\ \bottomrule
\end{tabular}%
\setlength\extrarowheight{0pt}
\label{tab:mlp_kan_comparison}
\end{table}

Our selected ML pipelines were compared to the top-performing model from the previous study (SVM) \cite{pew2017turn}, considered as the baseline, to demonstrate improved performance measured by the macro-average F1-score. In addition, our ML pipelines were compared to existing ML pipelines developed by Auto-sklearn \cite{feurer-neurips15a}.
 Auto-sklearn leverages automated machine learning (AutoML) techniques \cite{he2021automl} to evaluate a wide range of ML models. It incorporates state-of-the-art methods, including meta-learning for smart initialization, Bayesian optimization for hyperparameter tuning, and ensemble methods to combine predictions from different models. The time budget for Auto-sklearn was set to five minutes, with three-fold cross-validation for optimization.

\subsection{Handling class imbalance}

The straight walking class occurs more frequently than the stance and swing phases during turning (Table \ref{tab:data-proportion}). This results in class imbalance issue. Model development considered a weighting method \cite{king2001logistic} in the cross-entropy loss function \cite{mohasel2025micronas} of developed models to assign higher importance to the minority classes to address class imbalance:

\begin{equation}
\mathbf{w}_k = \frac{n}{\#C \times n_k},
\label{eq:class_weight}
\end{equation}
where $\#C$ is the total number of classes, $n$ is the total number of windows, and $n_k$ is the number of windows belonging to class $k$. This weighting scheme ensures that classes with fewer samples receive a higher weight, balancing the effect of all classes in the training.

\begin{table}[ht]
\caption{Proportion of training and testing data for each subject across the three classes: Straight walking (SW), stance (ST), and swing (SP). The table highlights the percentage distribution of each class within the training and testing sets, revealing an imbalanced class, with certain classes overrepresented or underrepresented.}
\begin{tabularx}{\linewidth}{@{}XXXXXXX@{}}
\toprule
\textbf{Subject} & \multicolumn{3}{c}{\textbf{Train} (\%)} & \multicolumn{3}{c}{\textbf{Test} (\%)} \\ \midrule
 & SW & ST & SP & SW & ST & SP \\ \cmidrule(lr){2-4} \cmidrule(lr){5-7}
A01 & $75.6$ & $15.1$ & $9.2$ & $73.4$ & $16.9$ & $9.7$ \\
A02 & $74.7$ & $17.2$ & $8.2$ & $73.4$ & $17.9$ & $8.7$ \\
A03 & $73.2$ & $17.7$ & $9.2$ & $71.6$ & $18.9$ & $9.5$ \\
A04 & $71.1$ & $18.3$ & $10.6$ & $70.8$ & $18.2$ & $11.0$ \\
A05 & $69.2$ & $20.8$ & $9.9$ & $69.0$ & $21.3$ & $9.7$ \\ \bottomrule
\end{tabularx}
\label{tab:data-proportion}
\end{table}

\subsection{Statistical tests}
We selected a Wilcoxon signed-rank test to evaluate our hypotheses ($p < 0.05$ determines significance). 

\textbf{Hypothesis 1} tests whether learnable activation functions lead to higher macro-averaged F1 scores than static ones.

\begin{itemize}
    \item \textbf{Null Hypothesis} ($H_0^1$): Macro-averaged F1 scores between KAN and MLP are similar. Formally,
    \[
    H_0^1 : \mathrm{F1}_{\mathrm{KAN}} = \mathrm{F1}_{\mathrm{MLP}}.
    \]

    \item \textbf{Alternative Hypothesis} ($H_1^1$): The KAN models yield a higher macro-averaged F1 score than the MLP models. Formally,
    \[
    H_1^1 : \mathrm{F1}_{\mathrm{KAN}} > \mathrm{F1}_{\mathrm{MLP}}.
    \]
\end{itemize}

The comparison between KAN and MLP evaluates the effectiveness of learnable activation functions in conventional methods. Similarly, FKAN and CNN were compared to assess the effectiveness of learnable activation functions in deep learning models equipped with time series feature extractors.

\begin{figure}[h!]
    \centering
    \includegraphics[width=\linewidth, height=10cm, keepaspectratio]{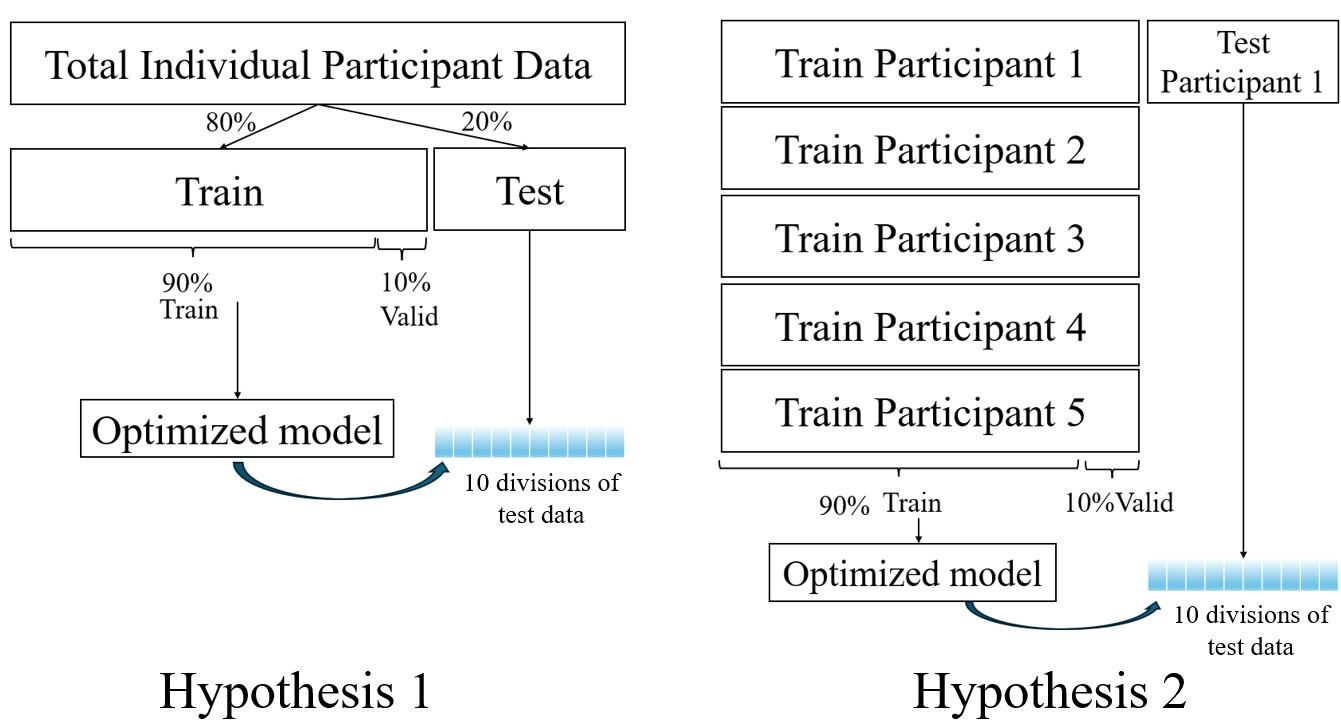}
    \caption{Model optimization and statistical analysis for hypothesis testing. Hypothesis 1 compares the 10 test data divisions between KAN and MLP (left panel). Hypothesis 2 compares the 10 test data divisions using pooled training from a single model (e.g., KAN) (right panel) with the 10 divisions using specific training from the same model (left panel).}
    \label{fig:subject_specific}
\end{figure}

Hypothesis 1 used the training data for each participant to train the model and tested its performance on that participant's test data (Fig. \ref{fig:subject_specific}, left). The macro-average F1-scores were compared across ten divisions of test data using stratified random sampling. A one-tailed (directional) Wilcoxon signed-rank test was used to assess the statistical significance of model performance differences for each participant, comparing MLP with KAN and CNN with FKAN ($p < 0.05$). 

To evaluate whether the hypothesis is supported across all participants, the average F-score across 10 divisions of test data was computed, yielding five paired samples. A one-tailed Bayesian paired t-test, which is suitable for small sample sizes, was then conducted to assess whether the results provided evidence in favor of the alternative hypothesis \cite{cleophas2018modern}.

\textbf{Hypothesis 2} evaluates the impact of training data on model performance. KAN models trained on individual-specific data are compared to those trained on pooled data.

\begin{itemize}
    \item \textbf{Null Hypothesis} ($H_0^2$): Macro-averaged F1 score between models trained on individual-specific data and those trained on pooled data are similar. Formally,
    \[
    H_0^2 : \mathrm{F1}_{\mathrm{individual}} = \mathrm{F1}_{\mathrm{pooled}}.
    \]

    \item \textbf{Alternative Hypothesis} ($H_1^2$): Models trained on individual-specific data achieve a higher macro-averaged F1 score compared to models trained on pooled data. Formally,
    \[
    H_1^2 : \mathrm{F1}_{\mathrm{individual}} > \mathrm{F1}_{\mathrm{pooled}}.
    \]
\end{itemize}

Hypothesis 2 utilized the training data from five participants for model training (pooled training). Then, the model was tested on ten divisions of a specific participant's test data (Fig. \ref{fig:subject_specific}, right). The macro-average F1-scores of the ten divisions were compared to those from models trained on participant-specific data (Fig. \ref{fig:subject_specific}, left) using a one-tailed Wilcoxon test. Ten divisions of the test data ensure direct comparability between Hypothesis 1 (subject-specific models) and Hypothesis 2 (pooled vs. individual training), as both hypotheses are evaluated on identical test partitions.

In addition to KAN and FKAN, we included MLP and CNN to test this hypothesis and compare the results with the previous study \cite{pew2017turn}. Finally, a one-tailed Bayesian paired t-test evaluates this hypothesis across all participants.

\subsection{Optimization}
To select the optimal architecture for each model, we used stratified random sampling to assign 10\% of the training data as validation data for hyperparameter tuning.  Figure \ref{fig:subject_specific} (left) illustrates the optimization process for Hypothesis 1, whereas Figure \ref{fig:subject_specific} (right) presents the corresponding optimization for Hypothesis 2.
Bayesian optimization with the macro-average F1-score was used for neural architecture search and hyperparameter tuning \cite{wu2019hyperparameter}, as summarized in Appendix Tables \ref{tab:nas-kan} and \ref{tab:nas-fkan} for MLP/KAN and CNN/FKAN, respectively. 

Model training and optimization was performed on a computer system equipped with an AMD EPYC 7252 processor, which has 8 cores and 16 threads, operating at a base clock speed of 3.1 GHz. The system includes 64 GiB of DDR4 memory. To expedite network training, it incorporates two NVIDIA A40 GPUs.

Additional analysis includes presenting the macro-average F1-scores for all models using subject-specific data to further quantify the differences between models for each participant, as well as confusion matrices to illustrate their relative performance. We also quantify the prediction time (on a single window) of MLP and KAN on the specified computer to approximate their difference in inference speed.

\section{Results}
\begin{figure*}[h!]
    \centering
    \includegraphics[width=0.95\linewidth]{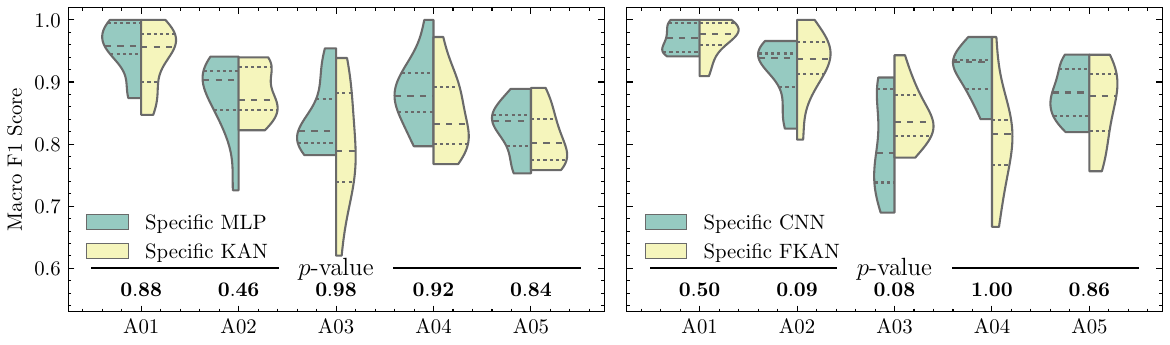}
    \caption{ Violin plots show the distribution of macro F1 scores for participant-specific models (Hypothesis 1). The left plot compares MLP and KAN, while the right compares CNN and FKAN. Horizontal bars indicate the median and interquartile range. p-values assess statistical significance for each participant.}
    \label{fig:hypothesis one_}
\end{figure*}

\textbf{Hypothesis 1:} For each participant, Wilcoxon signed-rank tests showed no significant difference between MLP and KAN for each individual ($p = 0.46$--$0.98$; Fig.~\ref{fig:hypothesis one_}, left).
 Similarly, comparisons between CNN and FKAN yielded $p$-values of 0.08--1.00 (Fig.~\ref{fig:hypothesis one_}, right), indicating no significant differences at the $\alpha = 0.05$ threshold.

The Bayesian paired t-test indicated no significant improvement in performance ($p >0.05$) when KAN was compared to MLP, or when FKAN was compared to CNN, across all participants (Table \ref{tab:model_comparison}).

\begin{table}[ht]
\centering
\caption{Statistical comparison between models using the paired t-test}
\label{tab:model_comparison}
\resizebox{0.9\linewidth}{!}{%
\begin{tabular}{@{}lll@{}}
\toprule
{\small \textbf{Hypothesis}}                   & \textbf{Model}                       & $p$-\textbf{value} \\ \midrule
\multirow{2}{*}{HP1} & Specific KAN - Specific MLP  & 0.965 \\
                     & Specific FKAN - Specific CNN & 0.701 \\ \midrule
\multirow{4}{*}{HP2} & Specific MLP - Pooled MLP    & 0.002 \\
                     & Specific KAN - Pooled KAN    & 0.015 \\
                     & Specific CNN - Pooled CNN    & 0.127 \\
                     & Specific FKAN - Pooled FKAN  & 0.833 \\ \bottomrule
\end{tabular}
}
\end{table}

\begin{figure*}[h!]
    \centering
    \includegraphics[width=0.95\linewidth]{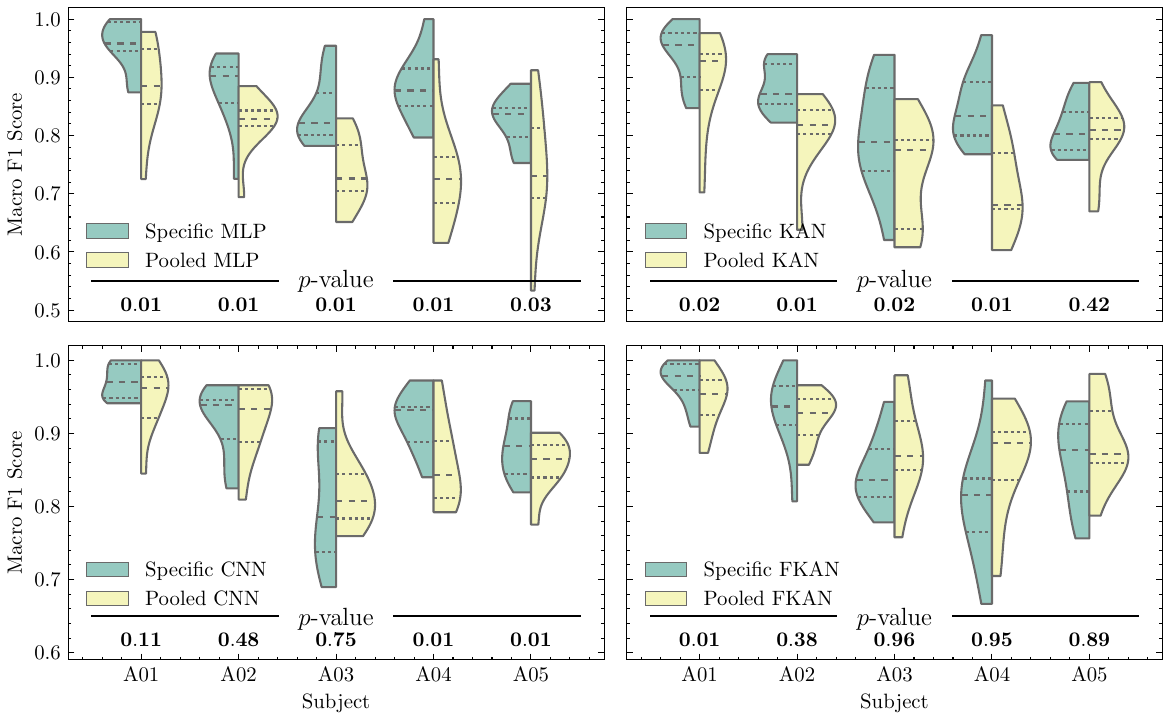}
    \caption{Comparison of Subject-Specific vs. Pooled Models Across Models (Hypothesis 2). Violin plots show the distribution of macro F1 scores for each participant across different models. Each subplot compares participant-specific training with pooled training (yellow) for MLP, KAN, CNN, and FKAN. Horizontal bars indicate the median and interquartile range. p-values assess statistical significance for each participant.}
    \label{fig:Pool_versus_specific_training}
\end{figure*}

\textbf{Hypothesis 2:} The Wilcoxon test indicated a statistically significant difference for MLP and KAN for specifc vs pooled training in each participant  ($p < 0.01$ and $p = 0.02$, respectively) (Fig.~\ref{fig:Pool_versus_specific_training}, top left). Similarly, a significant difference was observed for specific vs pooled training in each participant with KAN, except for participant 5 (Fig.~\ref{fig:Pool_versus_specific_training}, top right); however, no significant difference was observed for CNN (Fig.~\ref{fig:Pool_versus_specific_training}, bottom left) and FKAN (Fig.~\ref{fig:Pool_versus_specific_training}, bottom right) for the majority of participants ($p >0.05$).

The Bayesian paired t-test indicated a significant improvement in performance for MLP and KAN across all participants (\(p < 0.05\)). However, no evidence of improvement was observed for CNN and FKAN (Table~\ref{tab:model_comparison}).

\begin{table}[ht]
    \caption{Comparison of the macro average F1-scores for MLP, KAN, CNN, FKAN, Auto-sklearn, and the baseline model (SVM) \cite{pew2017turn} across individual subjects and the pooled dataset. The highest value for each individual is shown in bold, and the second-highest value is underlined.}
    \label{fig:compare-methods}
    \centering
\begin{tabular}{@{}ccccccc@{}}
\toprule
\textbf{Subject} & \textbf{MLP}   & \textbf{KAN}   & \textbf{CNN}   & \textbf{FKAN}  & \textbf{Auto-sklearn} & \textbf{SVM}\\ \midrule
A01     & 93.83 & 91.13 & 95.43 & \textbf{97.75} & \underline{96.95}  & 93.42    \\
A02     & 89.11 & 87.35 & \textbf{91.24} & \underline{90.83} & 87.53  & 90.77    \\
A03     & 83.27 & 80.09 & \textbf{85.61} & 85.50 & \underline{85.59}  & 80.67    \\
A04     & 83.09 & 84.83 & \underline{86.99} & 84.66 & \textbf{87.91}  & 86.93    \\
A05     & 82.13 & 80.75 & \underline{89.48} & 85.83 & 80.18  & \textbf{90.50}    \\
Pooled  & 83.08 & 84.44 & \underline{88.89} & \textbf{90.08} & 83.12  & 83.32    \\ \bottomrule
\end{tabular}
\end{table}

FKAN achieved the highest F1-score for participant A01 in both the individual (98\%) and the pooled datasets (90\%) (Table ~\ref{fig:compare-methods}). For participant A02, CNN outperformed other models, achieving an F1-score of 91\%. For participant A03, CNN, FKAN, and Auto-sklearn all achieved an F1-score of 85\%. The baseline SVM model \cite{pew2017turn} remained the best-performing model for participant A05, with a score of 90\%. The confusion matrices for all models are presented in Table~\ref{tab:confusion-matrices} in the Appendix. Additionally, a comparative analysis of the inference speeds of KAN and MLP is provided in Table~\ref{tab:time} in the Appendix.

\section{Discussion}

The primary objective of this study was to show improved control for prosthetic devices like the VSTA using machine learning with learnable activation functions, specifically, KAN and FKAN models. While FKAN achieved increased F1-score performance for 3 of 5 participants in comparison to the original SVM model, the learnable activation function failed consistently to outperform other techniques with static activation functions. Our methods outperformed Auto-sklearn in 4 of 5 participants. We set a high baseline by selecting Auto-sklearn; Auto-sklearn automatically develops several ML pipelines using a wide range of classifiers, with the final model often being an ensemble of the top-performing pipelines. While Auto-sklearn is competitive in achieving high macro-averaged F1 score performance, it may not be feasible for deployment on the VSTA controller, as the final ensemble (often a weighted combination of models) can be computationally more expensive than our single optimized model and may violate the constraint of real-time prediction.

\textbf{Hypothesis 1:} We hypothesized that learnable activation functions (specifically in KAN and FKAN) could outperform MLP and CNN, respectively, with static activations for the VSTA controller, however, the results rejected this hypothesis. One explanation may be related to the classification task, which involved only three labels: straight walking, stance, and swing phase. Static activation functions were likely sufficient for this level of complexity, whereas the additional parameters introduced by B-spline learnable activation functions may have led to overfitting, ultimately reducing performance.

Another reason for the rejection of Hypothesis 1 is the optimization of static activation functions (Tanh, ReLU, and SiLU) in models competing against learnable activation functions.
The optimization provides greater flexibility to the static activation functions, making it more challenging for the learnable activation functions to achieve superior performance, ultimately leading to the rejection of our hypothesis. Learnable activation functions could demonstrate their full potential when the classification task includes more actions \cite{liu2024initial}.

\textbf{Hypothesis 2:} We hypothesized that models trained on individual-specific data would achieve higher performance (macro average F1 score) for the VSTA controller for each participant compared to models trained on pooled data from multiple participants and tested on a specific participant. 

While MLP and KAN support this hypothesis and previous findings \cite{pew2017turn}, CNN and FKAN do not. Both CNN and FKAN use filters to first extract temporal, time-domain features from windows of IMU data and then send the extracted features to fully connected layers. CNN uses fixed activation functions at the nodes, while FKAN uses fractional Jacobi functions as a learnable activation function that is fixed at nodes. Both CNN and FKAN use weight sharing, which reduces the number of trainable parameters. However, the total number of parameters in these models (CNN/FKAN) is greater than our conventional models (MLP/KAN).

CNN and FKAN are DL models. DL performance improves with more diverse data \cite{bengio2017deep}, as it can better optimize the parameters responsible for feature extraction. However, the performance of conventional methods does not necessarily improve with increased data diversity. For instance, in SVM, only the support vectors, a small subset of the data, affect performance. Similarly, in KNN, only the K closest data points to the test sample determine the outcome, which is also a limited subset.
Therefore, individualized control can be more effective for conventional methods, as exposure to data from different gait patterns may mislead the model rather than enhance its performance.

Our developed methods (CNN and FKAN) outperform the top performing models in the previous study \cite{pew2017turn} (Table \ref{fig:compare-methods}) by utilizing data windows for feature extraction, incorporating a weighting method to handle class imbalance,  and considering the macro-average F1-score as the optimization metric, which is more suitable than accuracy for class imbalance.

The top performing model (SVM) in the previous study attained a higher F1-score than our developed models only for participant 5, which can be explained by the No Free Lunch theorem \cite{wolpert1997no}. This theorem states that no single model performs best for all possible problems, as a model's effectiveness depends on the specific characteristics of the dataset. For participant 5, SVM achieved the highest F1-score \cite{pew2017turn}, indicating that the different classes in participant 5's gait can be better classified by hyperplanes and the misclassification risk minimization mechanism in SVM rather than by prediction based on time-domain features and learnable activation functions in FKAN.

The model with the highest macro average F1 score should be selected for a specific participant if it meets the prediction constraint of VSTA. Utilizing the learnable activation function (B-Spline) resulted in increased inference time, approximately 1000 times greater than MLP (Table \ref{tab:time} in the Appendix). Since KAN is a recently developed method, few studies have focused on improving its inference speed \cite{lai2024efficient}, and more research is needed in this area.

The current results were obtained using a single IMU sensor with five participants. Collecting data from more participants or incorporating additional sensors could potentially improve these results.
A missed recognition may result in fall and injury when using a powered prosthetic leg. However, varying stiffness levels in VSTA is not expected to adversely affect mobility at self-selected speeds. This suggests that a missed prediction would lead to a reduced benefit rather than an injury, supporting the suitability of our models, as their performance improved compared to previous results \cite{pew2017turn}.

This study included only a small number of participants ($n = 5$). While the results offer some early insights, they should be interpreted with caution. In addition, we reported the inference time on a personal computer; future studies should report inference time when the model is deployed on the intended microcontroller.

\section{Conclusion}

This study presents one of the first investigations into the use of Kolmogorov-Arnold networks and their deep learning variant, FKAN, for intent recognition in lower-limb prosthesis control. Notably, FKAN achieved the best performance for one participant, and CNNs emerged as the top-performing model for most others, outperforming Auto-sklearn and SVM.
While learnable activation functions did not significantly outperform traditional static functions in this context, our findings highlight the importance of matching model complexity to the task and data characteristics. The relatively simple classification problem, along with a small dataset, may have limited the advantages of KANs' added flexibility. 
Our results reaffirm that subject-specific training remains a critical factor for achieving high classification performance in conventional models. However, the results also indicate that deep learning achieves similar performance using pooled data.

\section{Acknowledgment}
 Computational efforts were performed on the Tempest High Performance Computing System, operated and supported by University Information Technology Research Cyberinfrastructure at Montana State University.

\bibliographystyle{IEEEtran}
\bibliography{paper}

\begin{thebibliography}{10}
\providecommand{\url}[1]{#1}
\csname url@samestyle\endcsname
\providecommand{\newblock}{\relax}
\providecommand{\bibinfo}[2]{#2}
\providecommand{\BIBentrySTDinterwordspacing}{\spaceskip=0pt\relax}
\providecommand{\BIBentryALTinterwordstretchfactor}{4}
\providecommand{\BIBentryALTinterwordspacing}{\spaceskip=\fontdimen2\font plus
\BIBentryALTinterwordstretchfactor\fontdimen3\font minus \fontdimen4\font\relax}
\providecommand{\BIBforeignlanguage}[2]{{%
\expandafter\ifx\csname l@#1\endcsname\relax
\typeout{** WARNING: IEEEtran.bst: No hyphenation pattern has been}%
\typeout{** loaded for the language `#1'. Using the pattern for}%
\typeout{** the default language instead.}%
\else
\language=\csname l@#1\endcsname
\fi
#2}}
\providecommand{\BIBdecl}{\relax}
\BIBdecl

\bibitem{lawson2014robotic}
B.~E. Lawson, J.~Mitchell, D.~Truex, A.~Shultz, E.~Ledoux, and M.~Goldfarb, ``A robotic leg prosthesis: Design, control, and implementation,'' \emph{IEEE Robotics \& Automation Magazine}, vol.~21, no.~4, pp. 70--81, 2014.

\bibitem{hafner2002transtibial}
B.~J. Hafner, J.~E. Sanders, J.~M. Czerniecki, and J.~Fergason, ``Transtibial energy-storage-and-return prosthetic devices: a review of energy concepts and a proposed nomenclature.'' \emph{Journal of Rehabilitation Research \& Development}, vol.~39, no.~1, 2002.

\bibitem{rietman2002gait}
J.~Rietman, K.~Postema, and J.~Geertzen, ``Gait analysis in prosthetics: opinions, ideas and conclusions,'' \emph{Prosthetics and orthotics international}, vol.~26, no.~1, pp. 50--57, 2002.

\bibitem{pew2015design}
C.~Pew and G.~K. Klute, ``Design of lower limb prosthesis transverse plane adaptor with variable stiffness,'' \emph{Journal of Medical Devices}, vol.~9, no.~3, p. 035001, 2015.

\bibitem{pew2017pilot}
C.~Pew and G.~Klute, ``Pilot testing of a variable stiffness transverse plane adapter for lower limb amputees,'' \emph{Gait \& Posture}, vol.~51, pp. 104--108, 2017.

\bibitem{hudgins1993new}
B.~Hudgins, P.~Parker, and R.~N. Scott, ``A new strategy for multifunction myoelectric control,'' \emph{IEEE transactions on biomedical engineering}, vol.~40, no.~1, pp. 82--94, 1993.

\bibitem{chen2013locomotion}
B.~Chen, E.~Zheng, X.~Fan, T.~Liang, Q.~Wang, K.~Wei, and L.~Wang, ``Locomotion mode classification using a wearable capacitive sensing system,'' \emph{IEEE transactions on neural systems and rehabilitation engineering}, vol.~21, no.~5, pp. 744--755, 2013.

\bibitem{huang2008strategy}
H.~Huang, T.~A. Kuiken, R.~D. Lipschutz \emph{et~al.}, ``A strategy for identifying locomotion modes using surface electromyography,'' \emph{IEEE transactions on biomedical engineering}, vol.~56, no.~1, pp. 65--73, 2008.

\bibitem{pew2017turn}
C.~Pew and G.~K. Klute, ``Turn intent detection for control of a lower limb prosthesis,'' \emph{IEEE Transactions on Biomedical Engineering}, vol.~65, no.~4, pp. 789--796, 2017.

\bibitem{young2014analysis}
A.~Young, T.~Kuiken, and L.~Hargrove, ``Analysis of using emg and mechanical sensors to enhance intent recognition in powered lower limb prostheses,'' \emph{Journal of neural engineering}, vol.~11, no.~5, p. 056021, 2014.

\bibitem{young2013training}
A.~J. Young, A.~M. Simon, and L.~J. Hargrove, ``A training method for locomotion mode prediction using powered lower limb prostheses,'' \emph{IEEE Transactions on Neural Systems and Rehabilitation Engineering}, vol.~22, no.~3, pp. 671--677, 2013.

\bibitem{zhuang2019design}
W.~Zhuang, Y.~Chen, J.~Su, B.~Wang, and C.~Gao, ``Design of human activity recognition algorithms based on a single wearable imu sensor,'' \emph{International Journal of Sensor Networks}, vol.~30, no.~3, pp. 193--206, 2019.

\bibitem{yang2018wearable}
D.~Yang, J.~Huang, X.~Tu, G.~Ding, T.~Shen, and X.~Xiao, ``A wearable activity recognition device using air-pressure and imu sensors,'' \emph{IEEE access}, vol.~7, pp. 6611--6621, 2018.

\bibitem{ayman2019efficient}
A.~Ayman, O.~Attalah, and H.~Shaban, ``An efficient human activity recognition framework based on wearable imu wrist sensors,'' in \emph{2019 IEEE International Conference on Imaging Systems and Techniques (IST)}.\hskip 1em plus 0.5em minus 0.4em\relax IEEE, 2019, pp. 1--5.

\bibitem{golyski2017computational}
P.~R. Golyski and B.~D. Hendershot, ``A computational algorithm for classifying step and spin turns using pelvic center of mass trajectory and foot position,'' \emph{Journal of Biomechanics}, vol.~54, pp. 96--100, 2017.

\bibitem{novak2014toward}
D.~Novak, M.~Gor{\v{s}}i{\v{c}}, J.~Podobnik, and M.~Munih, ``Toward real-time automated detection of turns during gait using wearable inertial measurement units,'' \emph{Sensors}, vol.~14, no.~10, pp. 18\,800--18\,822, 2014.

\bibitem{kaur2019systematic}
H.~Kaur, H.~S. Pannu, and A.~K. Malhi, ``A systematic review on imbalanced data challenges in machine learning: Applications and solutions,'' \emph{ACM computing surveys (CSUR)}, vol.~52, no.~4, pp. 1--36, 2019.

\bibitem{mohasel2025robust}
S.~M. Mohasel and H.~Koosha, ``Robust support vector machines for imbalanced and noisy data via benders decomposition,'' \emph{arXiv preprint arXiv:2503.14873}, 2025.

\bibitem{ali2019imbalance}
H.~Ali, M.~M. Salleh, R.~Saedudin, K.~Hussain, and M.~F. Mushtaq, ``Imbalance class problems in data mining: A review,'' \emph{Indonesian Journal of Electrical Engineering and Computer Science}, vol.~14, no.~3, pp. 1560--1571, 2019.

\bibitem{johnson2019survey}
J.~M. Johnson and T.~M. Khoshgoftaar, ``Survey on deep learning with class imbalance,'' \emph{Journal of big data}, vol.~6, no.~1, pp. 1--54, 2019.

\bibitem{wang2019deep}
J.~Wang, Y.~Chen, S.~Hao, X.~Peng, and L.~Hu, ``Deep learning for sensor-based activity recognition: A survey,'' \emph{Pattern recognition letters}, vol. 119, pp. 3--11, 2019.

\bibitem{hargrove2013non}
L.~J. Hargrove, A.~M. Simon, R.~Lipschutz, S.~B. Finucane, and T.~A. Kuiken, ``Non-weight-bearing neural control of a powered transfemoral prosthesis,'' \emph{Journal of neuroengineering and rehabilitation}, vol.~10, pp. 1--11, 2013.

\bibitem{ha2010volitional}
K.~H. Ha, H.~A. Varol, and M.~Goldfarb, ``Volitional control of a prosthetic knee using surface electromyography,'' \emph{IEEE Transactions on Biomedical Engineering}, vol.~58, no.~1, pp. 144--151, 2010.

\bibitem{huang2011continuous}
H.~Huang, F.~Zhang, L.~J. Hargrove, Z.~Dou, D.~R. Rogers, and K.~B. Englehart, ``Continuous locomotion-mode identification for prosthetic legs based on neuromuscular--mechanical fusion,'' \emph{IEEE Transactions on Biomedical Engineering}, vol.~58, no.~10, pp. 2867--2875, 2011.

\bibitem{hu2018fusion}
B.~Hu, E.~Rouse, and L.~Hargrove, ``Fusion of bilateral lower-limb neuromechanical signals improves prediction of locomotor activities,'' \emph{Frontiers in Robotics and AI}, vol.~5, p.~78, 2018.

\bibitem{bengio2013deep}
Y.~Bengio, ``Deep learning of representations: Looking forward,'' in \emph{International conference on statistical language and speech processing}.\hskip 1em plus 0.5em minus 0.4em\relax Springer, 2013, pp. 1--37.

\bibitem{yang2015deep}
J.~Yang, M.~N. Nguyen, P.~P. San, X.~Li, and S.~Krishnaswamy, ``Deep convolutional neural networks on multichannel time series for human activity recognition.'' in \emph{Ijcai}, vol.~15.\hskip 1em plus 0.5em minus 0.4em\relax Buenos Aires, Argentina, 2015, pp. 3995--4001.

\bibitem{lu2020deep}
Z.~Lu, A.~Narayan, and H.~Yu, ``A deep learning based end-to-end locomotion mode detection method for lower limb wearable robot control,'' in \emph{2020 IEEE/RSJ International Conference on Intelligent Robots and Systems (IROS)}.\hskip 1em plus 0.5em minus 0.4em\relax IEEE, 2020, pp. 4091--4097.

\bibitem{su2019cnn}
B.-Y. Su, J.~Wang, S.-Q. Liu, M.~Sheng, J.~Jiang, and K.~Xiang, ``A cnn-based method for intent recognition using inertial measurement units and intelligent lower limb prosthesis,'' \emph{IEEE Transactions on Neural Systems and Rehabilitation Engineering}, vol.~27, no.~5, pp. 1032--1042, 2019.

\bibitem{nazari2022comparison}
F.~Nazari, N.~Mohajer, D.~Nahavandi, A.~Khosravi, and S.~Nahavandi, ``Comparison study of inertial sensor signal combination for human activity recognition based on convolutional neural networks,'' in \emph{2022 15th International Conference on Human System Interaction (HSI)}.\hskip 1em plus 0.5em minus 0.4em\relax IEEE, 2022, pp. 1--6.

\bibitem{liu2024kan}
Z.~Liu, Y.~Wang, S.~Vaidya, F.~Ruehle, J.~Halverson, M.~Solja{\v{c}}i{\'c}, T.~Y. Hou, and M.~Tegmark, ``Kan: Kolmogorov-arnold networks,'' \emph{arXiv preprint arXiv:2404.19756}, 2024.

\bibitem{liu2024kan2}
Z.~Liu, P.~Ma, Y.~Wang, W.~Matusik, and M.~Tegmark, ``Kan 2.0: Kolmogorov-arnold networks meet science,'' \emph{arXiv preprint arXiv:2408.10205}, 2024.

\bibitem{aghaei2024fkan}
A.~Afzal~Aghaei, ``fkan: Fractional kolmogorov--arnold networks with trainable jacobi basis functions,'' \emph{Neurocomputing}, vol. 623, p. 129414, 2025.

\bibitem{al2024tcnn}
M.~A. Al-Qaness and S.~Ni, ``Tcnn-kan: Optimized cnn by kolmogorov-arnold network and pruning techniques for semg gesture recognition,'' \emph{IEEE Journal of Biomedical and Health Informatics}, 2024.

\bibitem{vaca2024kolmogorov}
C.~J. Vaca-Rubio, L.~Blanco, R.~Pereira, and M.~Caus, ``Kolmogorov-arnold networks (kans) for time series analysis,'' \emph{arXiv preprint arXiv:2405.08790}, 2024.

\bibitem{livieris2024c}
I.~E. Livieris, ``C-kan: A new approach for integrating convolutional layers with kolmogorov--arnold networks for time-series forecasting,'' \emph{Mathematics}, vol.~12, no.~19, p. 3022, 2024.

\bibitem{liu2024initial}
M.~Liu, D.~Gei{\ss}ler, D.~Nshimyimana, S.~Bian, B.~Zhou, and P.~Lukowicz, ``Initial investigation of kolmogorov-arnold networks (kans) as feature extractors for imu based human activity recognition,'' in \emph{Companion of the 2024 on ACM International Joint Conference on Pervasive and Ubiquitous Computing}, 2024, pp. 500--506.

\bibitem{yang2024adaptive}
G.~Yang, J.~Heo, and B.~B. Kang, ``Adaptive vision-based gait environment classification for soft ankle exoskeleton,'' in \emph{Actuators}, vol.~13, no.~11.\hskip 1em plus 0.5em minus 0.4em\relax MDPI, 2024, p. 428.

\bibitem{bengio2017deep}
Y.~Bengio, I.~Goodfellow, and A.~Courville, \emph{Deep learning}.\hskip 1em plus 0.5em minus 0.4em\relax MIT press Cambridge, MA, USA, 2017, vol.~1.

\bibitem{elbaz2018gait}
A.~Elbaz, F.~Artaud, A.~Dugravot, C.~Tzourio, and A.~Singh-Manoux, ``The gait speed advantage of taller stature is lost with age,'' \emph{Scientific reports}, vol.~8, no.~1, p. 1485, 2018.

\bibitem{kim2021effects}
D.~Kim, C.~L. Lewis, and S.~V. Gill, ``Effects of obesity and foot arch height on gait mechanics: A cross-sectional study,'' \emph{Plos one}, vol.~16, no.~11, p. e0260398, 2021.

\bibitem{glaister2007video}
B.~C. Glaister, G.~C. Bernatz, G.~K. Klute, and M.~S. Orendurff, ``Video task analysis of turning during activities of daily living,'' \emph{Gait \& posture}, vol.~25, no.~2, pp. 289--294, 2007.

\bibitem{deathe2005test}
A.~B. Deathe and W.~C. Miller, ``The l test of functional mobility: measurement properties of a modified version of the timed “up \& go” test designed for people with lower-limb amputations,'' \emph{Physical therapy}, vol.~85, no.~7, pp. 626--635, 2005.

\bibitem{joseph2022split}
V.~R. Joseph and A.~Vakayil, ``Split: An optimal method for data splitting,'' \emph{Technometrics}, vol.~64, no.~2, pp. 166--176, 2022.

\bibitem{banos2014window}
O.~Banos, J.-M. Galvez, M.~Damas, H.~Pomares, and I.~Rojas, ``Window size impact in human activity recognition,'' \emph{Sensors}, vol.~14, no.~4, pp. 6474--6499, 2014.

\bibitem{taud2017multilayer}
H.~Taud and J.-F. Mas, ``Multilayer perceptron (mlp),'' in \emph{Geomatic approaches for modeling land change scenarios}.\hskip 1em plus 0.5em minus 0.4em\relax Springer, 2017, pp. 451--455.

\bibitem{feurer-neurips15a}
M.~Feurer, A.~Klein, K.~Eggensperger, J.~Springenberg, M.~Blum, and F.~Hutter, ``Efficient and robust automated machine learning,'' in \emph{Advances in Neural Information Processing Systems 28 (2015)}, 2015, pp. 2962--2970.

\bibitem{he2021automl}
X.~He, K.~Zhao, and X.~Chu, ``Automl: A survey of the state-of-the-art,'' \emph{Knowledge-based systems}, vol. 212, p. 106622, 2021.

\bibitem{king2001logistic}
G.~King and L.~Zeng, ``Logistic regression in rare events data,'' \emph{Political analysis}, vol.~9, no.~2, pp. 137--163, 2001.

\bibitem{mohasel2025micronas}
S.~M. Mohasel, J.~Sheppard, L.~K. Molina, R.~R. Neptune, S.~R. Wurdeman, and C.~A. Pew, ``Micronas: An automated framework for developing a fall detection system,'' \emph{arXiv preprint arXiv:2504.07397}, 2025.

\bibitem{cleophas2018modern}
T.~J. Cleophas, A.~H. Zwinderman \emph{et~al.}, ``Modern bayesian statistics in clinical research,'' Springer, Tech. Rep., 2018.

\bibitem{wu2019hyperparameter}
J.~Wu, X.-Y. Chen, H.~Zhang, L.-D. Xiong, H.~Lei, and S.-H. Deng, ``Hyperparameter optimization for machine learning models based on bayesian optimization,'' \emph{Journal of Electronic Science and Technology}, vol.~17, no.~1, pp. 26--40, 2019.

\bibitem{wolpert1997no}
D.~H. Wolpert and W.~G. Macready, ``No free lunch theorems for optimization,'' \emph{IEEE transactions on evolutionary computation}, vol.~1, no.~1, pp. 67--82, 1997.

\bibitem{lai2024efficient}
Z.~Lai, Y.~Zhou, P.~Zheng, and L.~Chen, ``Efficient privacy-preserving kan inference using homomorphic encryption,'' \emph{arXiv preprint arXiv:2409.07751}, 2024.

\end{thebibliography}

\section{Appendix}

\begin{table}[ht]
\centering
\caption{Hyperparameter ranges for MLP and KAN}
\label{tab:nas-kan}
\begin{tabularx}{\linewidth}{@{}lll@{}}
\toprule
\textbf{Network Type} & \textbf{Parameter} & \textbf{Range} \\ \midrule
\multirow{3}{*}{\textbf{Both}} & Layers & Up to 5 \\
 & Neurons per Layer & 5-100 \\
 & Regularization & $[10^{-5}, 10^{-1}]$ \\
\multirow{1}{*}{\textbf{MLP}} & Activation Functions & Tanh, ReLU, SiLU \\
\multirow{2}{*}{\textbf{KAN}} & $k$ & $[1, 5]$ \\
 & Grid Size & $[1, 15]$ \\ \bottomrule
\end{tabularx}%
\end{table}

\begin{table}[ht]
    \centering
    \caption{CNN and FKAN Architecture for NAS}
    \label{tab:nas-fkan}
    \begin{tabularx}{\linewidth}{@{}ll@{}}
        \toprule
        \textbf{Parameter}                     & \textbf{Range}                                      \\ \midrule
        Input Size                           & $10 \times 6, 20 \times 6, 30 \times 6$                                        \\
        Convolutional Layers                 & Up to 6                                             \\
        Padding                              & Valid, Same                                                \\
        Number of Filters                    & 5 to 200                                            \\
        Kernel Sizes                         & 7 to 15                                             \\
        Activation Functions                  & ReLU, Tanh, FKAN 1-6                                \\
        Max Pooling Layers                   & After each Conv layer, Pool size up to 3                               \\
        Dropout Rate                         & 0.2 to 0.8                                          \\
        Global Average Pooling               & Optional before classifier                           \\
        Classifier                           & MLP with up to 3 layers                             \\
        Neurons per Layer                    & 10 to 500                                           \\
        Classifier Activations               & ReLU, Tanh                                         \\
        Optimizer                            & Adam                                                \\
        Learning Rate                        & $10^{-4}$ to $10^{-2}$                              \\ \bottomrule
    \end{tabularx}
\end{table}

\begin{table*}[t]
\caption{This table compares the training and inference times for optimized KAN and MLP architectures after hyperparameter optimization. It includes the number of training and testing samples for each subject, along with various architectural configurations and network hyperparameters, excluding the regularization parameter as it does not influence these times. The table details both total training durations and single-sample inference times for each model. Both KAN and MLP models underwent full-batch training for 50 epochs on 2-core CPU.}
\centering
\resizebox{\textwidth}{!}{%
\begin{tabular}{@{}cccccccccccc@{}}
\toprule
Subject & \# Train & \# Test & \multicolumn{5}{c}{KAN} & \multicolumn{4}{c}{MLP} \\ \midrule
 &  &  & Architecture & \begin{tabular}[c]{@{}c@{}}Grid\\ Size\end{tabular} & k & \begin{tabular}[c]{@{}c@{}}Total Data\\ Train Time\\ (s)\end{tabular} & \begin{tabular}[c]{@{}c@{}}Total Data\\ Inference\\ Time (s)\end{tabular} & Architecture & Activation & \begin{tabular}[c]{@{}c@{}}Total\\ Train\\ Time (s)\end{tabular} & \begin{tabular}[c]{@{}c@{}}Total Data\\ Inference\\ Time (s)\end{tabular} \\ \cmidrule(lr){4-8} \cmidrule(lr){9-12}
A01 & 1758 & 516 & {[}80{]} & 9 & 1 & 182 & 0.2 & {[}80, 35, 80, 45{]} & Tanh & 52 & 0.02 \\
A02 & 2344 & 607 & {[}40, 55, 75{]} & 5 & 2 & 2580 & 1.2 & {[}65, 50, 45{]} & SiLU & 43 & 0.01 \\
A03 & 2007 & 515 & {[}55, 65{]} & 5 & 4 & 1850 & 0.8 & {[}80, 40, 70, 55{]} & Tanh & 80 & 0.02 \\
A04 & 1802 & 436 & {[}45{]} & 3 & 2 & 350 & 0.1 & {[}30{]} & SiLU & 18 & 0.01 \\
A05 & 2233 & 568 & {[}80, 75{]} & 3 & 4 & 2390 & 1.2 & {[}90, 65, 25, 10, 30{]} & ReLU & 71 & 0.01 \\
Pooled & 10152 & 2648 & {[}100, 65, 90{]} & 3 & 2 & 12936 & 2.9 & {[}50, 40, 55, 90, 90{]} & Tanh & 144 & 0.03 \\ \bottomrule
\end{tabular}
}
\label{tab:time}
\end{table*}

\begin{table*}[t]
\caption{Normalized 3x3 confusion matrices for subjects A01 to A05 and the pooled dataset across five machine learning methods: MLP, KAN, CNN, FKAN, and Auto-sklearn. Each confusion matrix represents classification performance on three classes: SW, ST, and SP, with true labels in rows and predicted labels in columns. The values in each matrix are normalized across rows to reflect class-specific prediction accuracy.}
\centering
\renewcommand{\arraystretch}{1.2}
\begin{tabular}{@{}ccccc:ccc:ccc:ccc:ccc@{}}
    \toprule
    Subject & Label & \multicolumn{3}{c}{MLP} & \multicolumn{3}{c}{KAN} & \multicolumn{3}{c}{CNN} & \multicolumn{3}{c}{FKAN}  & \multicolumn{3}{c}{Auto-sklearn} \\
    \midrule
        &  & SW & ST & SP & SW & ST & SP & SW & ST & SP & SW & ST & SP  & SW & ST & SP \\ \cmidrule(lr){3-5} \cmidrule(lr){6-8} \cmidrule(lr){9-11} \cmidrule(lr){12-14} \cmidrule(lr){15-17}
    \multirow{3}{*}{A01} & SW & \textbf{98.2} & 1.3 & 0.5 & \textbf{95.8} & 1.8 & 2.4 & \textbf{97.6} & 2.1 & 0.3 & \textbf{99.5} & 0.5 & 0 & \textbf{98.9} & 0.8 & 0.3 \\
     & ST &                     13.6 & \textbf{85.2} & 1.1 & 5.7 & \textbf{93.2} & 1.1 & 4.5 & \textbf{93.2} & 2.3 & 2.3 & \textbf{96.6} & 1.1 & 5.7 & \textbf{94.3} & 0 \\
     & SP &                     4.1 & 0 & \textbf{95.9} & 12.2 & 0 & \textbf{87.8} & 2.0 & 0 & \textbf{98.0} & 4.1 & 0 & \textbf{95.9} & 4.1 & 0 & \textbf{95.9} \\
     \spacedhdashline
    \multirow{3}{*}{A02} & SW & \textbf{95.1} & 1.8 & 3.1 & \textbf{92.2} & 2.5 & 5.4 & \textbf{96.0} & 1.6 & 2.5 & \textbf{96.2} & 1.1 & 2.7 & \textbf{96.4} & 1.6 & 2.0 \\
     & ST &                     4.7 & \textbf{95.3} & 0 & 3.8 & \textbf{96.2} & 0 & 5.7 & \textbf{94.3} & 0 & 15.1 & \textbf{84.9} & 0 & 8.5 & \textbf{91.5} & 0 \\
     & SP &                     15.1 & 3.8 & \textbf{81.1} & 9.4 & 3.8 & \textbf{86.8} & 11.3 & 1.9 & \textbf{86.8} & 5.7 & 0 & \textbf{94.3} & 30.2 & 0 & \textbf{69.8} \\
    \spacedhdashline
    \multirow{3}{*}{A03} & SW & \textbf{92.4} & 4.3 & 3.3 & \textbf{90.8} & 4.9 & 4.3 & \textbf{93.5} & 3.3 & 3.3 & \textbf{96.2} & 2.2 & 1.6 & \textbf{98.9} & 0.8 & 0.3 \\
     & ST &                     18.4 & \textbf{79.6} & 2.0 & 19.4 & \textbf{76.5} & 4.1 & 20.4 & \textbf{76.5} & 3.1 & 24.5 & \textbf{73.5} & 2.0 & 22.4 & \textbf{75.5} & 2.0 \\
     & SP &                     18.4 & 2.0 & \textbf{79.6} & 18.4 & 4.1 & \textbf{77.6} & 6.1 & 2.0 & \textbf{91.8} & 14.3 & 4.1 & \textbf{81.6} & 32.7 & 0 & \textbf{67.3} \\
    \spacedhdashline
    \multirow{3}{*}{A04} & SW & \textbf{93.9} & 4.2 & 1.9 & \textbf{93.6} & 4.5 & 1.9 & \textbf{91.3} & 2.9 & 5.8 & \textbf{91.3} & 5.5 & 3.2 & \textbf{96.8} & 1.9 & 1.3 \\
     & ST &                     12.3 & \textbf{85.2} & 2.5 & 8.6 & \textbf{88.9} & 2.5 & 8.6 & \textbf{90.1} & 1.2 & 6.2 & \textbf{92.6} & 1.2 & 16.0 & \textbf{82.7} & 1.2 \\
     & SP &                     34.1 & 0 & \textbf{65.9} & 29.5 & 0 & \textbf{70.5} & 9.1 & 0 & \textbf{90.9} & 20.5 & 4.5 & \textbf{75.0} & 22.7 & 0 & \textbf{77.3} \\
    \spacedhdashline
    \multirow{3}{*}{A05} & SW & \textbf{91.9} & 4.1 & 4.1 & \textbf{90.4} & 6.1 & 3.6 & \textbf{94.2} & 3.8 & 2.0 & \textbf{89.8} & 5.6 & 4.6 & \textbf{95.9} & 3.6 & 0.5 \\
     & ST &                     18.3 & \textbf{80.0} & 1.7 & 22.5 & \textbf{75.8} & 1.7 & 15.8 & \textbf{82.5} & 1.7 & 10.0 & \textbf{88.3} & 1.7 & 25.8 & \textbf{73.3} & 0.8 \\
     & SP &                     20.4 & 3.7 & \textbf{75.9} & 18.5 & 3.7 & \textbf{77.8} & 1.9 & 1.9 & \textbf{96.3} & 7.4 & 1.9 & \textbf{90.7} & 37.0 & 5.6 & \textbf{57.4} \\
    \spacedhdashline
    \multirow{3}{*}{Pooled} & SW & \textbf{92.0} & 4.8 & 3.2 & \textbf{92.0} & 4.5 & 3.5 & \textbf{94.1} & 3.4 & 2.6 & \textbf{94.6} & 3.4 & 2.0 & \textbf{96.5} & 2.2 & 1.4 \\
     & ST &                        12.7 & \textbf{84.9} & 2.4 & 12.0 & \textbf{86.5} & 1.4 & 12.0 & \textbf{86.9} & 1.0 & 8.8 & \textbf{90.2} & 1.0 & 23.3 & \textbf{76.1} & 0.6 \\
     & SP &                        20.6 & 4.0 & \textbf{75.4} & 17.1 & 3.6 & \textbf{79.4} & 8.3 & 1.6 & \textbf{90.1} & 9.5 & 1.6 & \textbf{88.9} & 30.2 & 4.0 & \textbf{65.9} \\ \bottomrule
\end{tabular}
\label{tab:confusion-matrices}
\end{table*}

\end{document}